\def\year{2022}\relax
\documentclass[letterpaper]{article} 
\usepackage{aaai22}  
\usepackage{times}  
\usepackage{helvet}  
\usepackage{courier}  
\usepackage[hyphens]{url}  
\usepackage{graphicx} 
\usepackage{natbib}  
\usepackage{caption} 
\DeclareCaptionStyle{ruled}{labelfont=normalfont,labelsep=colon,strut=off} 
\usepackage{float}
\usepackage{algorithm}  
\usepackage{algorithmicx}  
\usepackage{algpseudocode}  
\usepackage{amsmath}
\usepackage{float}
\usepackage{amssymb}
\usepackage{amsmath}

\usepackage{microtype}
\usepackage{graphicx}
\usepackage{subfigure}
\usepackage{booktabs} 
\usepackage{amsfonts}
\usepackage{amssymb}
\usepackage{amsmath}
\usepackage{bm}

\usepackage{color, colortbl}
\definecolor{LightCyan}{rgb}{0.88,1,1}
\definecolor{Gray}{gray}{0.9}

\usepackage{newfloat}
\usepackage{listings}
\floatstyle{ruled}
\newfloat{listing}{tb}{lst}{}
\floatname{listing}{Listing}
\title{Sparse-softmax: A Simpler and Faster \\ Alternative Softmax Transformation}

\author{
    	Shaoshi Sun\textsuperscript{1},
    Zhenyuan Zhang\textsuperscript{2},
    BoCheng	Huang\textsuperscript{3},
    	Pengbin Lei\textsuperscript{4}, \\ 
    Jianlin Su\textsuperscript{5}, 
    Shengfeng Pan\textsuperscript{5}, 
    Jiarun Cao\textsuperscript{6}
}
\affiliations{
    \textsuperscript{\rm 1}School of Computer Science and Informatics, Cardiff University, the United Kingdom	\\
    
    \textsuperscript{\rm 2}Department of Economics, Osaka City University, Japan \\
    \textsuperscript{\rm 3}School of Software Engineering, Beijing Jiaotong University, China \\ 
    \textsuperscript{\rm 4}College of Electronics and Information Engineering, Shenzhen University, China \\
    \textsuperscript{\rm 5}~Shenzhen Zhuiyi Technology Co., Ltd., China \\
    \textsuperscript{\rm 6}~Chinese Academy of Science, China \\

}

\usepackage{bibentry}

\usepackage{hyperref}
\hypersetup{
    colorlinks=false,
    linkcolor=blue,
    filecolor=magenta,      
    urlcolor=cyan,
    pdftitle={Overleaf Example},
    pdfpagemode=FullScreen,
    }

\begin{document}

\maketitle

\begin{abstract}
The softmax function is widely used in artiﬁcial neural networks for the multiclass classiﬁcation problems, where the softmax transformation enforces the output to be positive and sum to one, and the corresponding loss function allows to use maximum likelihood principle to optimize the model. However, softmax leaves a large margin for loss function to conduct optimizing operation when it comes to high-dimensional classification, which results in low-performance to some extent. In this paper, we provide an empirical study on a simple and concise softmax variant, namely sparse-softmax, to alleviate the problem that occurred in traditional softmax in terms of high-dimensional classification problems. We evaluate our approach in several interdisciplinary tasks, the experimental results show that sparse-softmax is simpler, faster, and produces better results than the baseline models.
\end{abstract}

The Softmax transformation is widely used in artiﬁcial neural networks for multi-class classification,multi-label classification and attention mechanisms where it typically appears as the last layer. However, when it comes to classification problems with high dimensional outputs(empirically more than 100 categories), the standard softmax and backpropagation do not take advantage of the sparsity of the categories and, as a result, softmax converges slowly on high-dimensional classification tasks. The softmax function has often been scrutinized in search of finding a better alternative to tackle the problem aforementioned. Specifically, the first direction is sampling methods approximations, which compute a fraction of the output's dimensions~\citep{gutmann2010noise,mnih2013learning,mikolov2013distributed,shrivastava2014asymmetric}. The second direction is modeling high dimensional classification as a hierarchical classification task, where it modifies the output softmax layer by introducing heuristical-based tree~\citep{mikolov2013distributed,morin2005hierarchical}.

Furthermore, ~\citep{vincent2015efficient} explore the spherical loss family where they propose an alternative softmax that has log-softmax loss as one of its members. ~\citep{de2015exploration} further work on this family of loss functions and propose log-Taylor softmax as a superior alternative than others, including original log-softmax loss. ~\citep{liu2016large} propose large-margin softmax (LM-softmax) that tries to increase inter-class separation and decrease intra-class separation. This approach is further investigated by ~\citep{liang2017soft}, where they propose soft-margin softmax (SM-softmax) that provides a ner control over the inter-class separation compared to LM-softmax. 

In this work, we propose a simple and scalable alternative softmax namely Sparse-softmax, which specifically takes an effect on the high-dimensional classification problems. We first describe precisely our approach and explain theoretically why it is effective. Then we evaluate our approach on the interdisciplinary tasks including three NLP tasks: text classification~\citep{yang2019xlnet,howard2018universal}, abstractive summarization~\citep{al2018hierarchical,nallapati2016abstractive}, question generation~\citep{kundu2018question,tay2018densely}, as well as an image classification in Cifar-100 dataset. Experimental results indicate that our approach outperforms the baseline models. Our contribution can be summarized as the following:
\begin{itemize}
\item We introduce a simple softmax alternative called \textbf{sparse-softmax}, and its corresponding loss function during training. 
\item We explain the reason in-depth why \textbf{sparse-softmax} advances normal softmax in high-dimensional classification problem. 
\item We design interdisciplinary experiments to exhaustively analyse our model, where the experimental result verifies our model effectiveness. 
\end{itemize}

\section{Methods}

In this section, we provide a brief overview of the softmax transformation and cross-entropy loss function in section~\ref{background}. Then, We propose our sparse-softmax and a modified loss function in section~\ref{sparse}, furthermore, we elaborate on the inner mechanism of the advance of our approach in high-dimensional classification problems. 
\subsection{Background}
\label{background}
\textbf{Definition. } We denote the $d$-dimensional simplex by ${\Delta}^d = \{ \bm{p} \in \mathbb{R}^d:\bm{p} \geqslant 0, \| \bm{p} \|_1  = 1 \}$, where the set of vectors represents probability distributions over $d$ categories. Empirically, we consider a task as a high-dimensional classification problem if $d \geqslant 100$. 

\textbf{Softmax. }We focus on the transformations that convert vectors in $\mathbb{R}^d$ to probability distributions in ${\Delta}^d$. One of the most well-studied one is the \textbf{softmax} function that converts a vector of weights to a posterior label probabilities. The softmax function is defined as following:
\begin{equation}
{p_i = Softmax(\bm{z})_i = \frac{e^{\bm{z}_i}}{\sum^d_{j=1} e^{\bm{z}_j}}}
\end{equation}
where the exponential function is executed on each element $\bm{z}_i$ of the input vector $\bm{z}$ and the output values are normalized by dividing by the sum of the entire exponentials. The normalization operation let each element in the output vector  $p_i$ sum up to 1. 

\textbf{Cross-entropy loss function. }To derive the loss function for the softmax function we start out from the likelihood function that a given set of parameters of the model can result in the prediction of the correct class of each input sample, as in the derivation for the logistic loss function. The maximization of this likelihood can be written as: 

\begin{equation}\label{equation2}
\mathop{\arg\max}\limits_{\theta} \mathcal{L}{(\theta|\bm{t},\bm{z})}
\end{equation}

Maximizing this likelihood can also be done by minimizing the negative log-likelihood:
\begin{equation}\label{equation3}
\mathcal{L}_{ce}= - \log p_t = \sum_i e^{\bm{z}_i} - \bm{z}_t 
\end{equation}
where $t$ denotes target category and $\bm{z}$ is the logits deriving from the output of softmax layer. 

\subsection{Sparse-softmax Algorithm}
\label{sparse}
\textbf{Approach. } A limitation of the conventional softmax function is that the resulting probability distribution always has full support across each $\bm{z}_i$, in another word, $softmax_i(\bm{z}) \ne 0$ for every $\bm{z}_i$.  
This is a weakness when it comes to high-dimensional classification problems where a sparse probability distribution is desired. 

In this paper, we propose an alternative transformation, which we call sparse-softmax, to tackle the limitation aforementioned. The idea of sparse-softmax is intuitive and concise: we manually set up a hyperparameter $k$, then we only select the maximum $k$ input values as a vector $\Omega_k \in \mathbb{R}^k $ to pass through the exponential normalized function, while others are masked as $0$:

\begin{equation}
{
    Sparse \, Softmax(\bm{z})_i = 
    \begin{cases}
     \frac{e^{\bm{z}_i}}{\sum_{j \in \Omega_k} e^{\bm{z}_j}} , & \bm{z}_i \in \Omega_k,\\
    0, & \bm{z}_i \notin \Omega_k
    \end{cases}
}
\end{equation}
where $\Omega_k$ is the set of top-k maximum indices of $\bm{z}_i$. Accordingly, the cross entropy loss function for sparse-softmax $\mathcal{L}_{sparse}$ is modified as:
\begin{equation}
{
\mathcal{L}_{sparse}= \log \sum_{i \in \Omega_k} e^{\bm{z}_i} - \bm{z}_t 
}
\end{equation}
where $\bm{z}_t$ is the logit of the targeted category. 

\textbf{Theoretical analysis}
In this part, we theoretically explain why our sparse-softmax is effective on high-dimensional classification tasks. As we elaborated in Equation~\ref{equation2} and ~\ref{equation3}, the conventional cross-entropy loss function can also be written as the following:

\begin{equation}
    {
    \mathcal{L}_{ce}= \log  (1 + \sum_{i \ne {t}} e^{\bm{z}_i-\bm{z}_t} )
}
\end{equation}
where $t$ is the targeted category, $\bm{z}$ is the logits. Presumably, we classify the current sample correctly, that is, $\bm{z}_{max} = \bm{z}_t$. Therefore, we can derive the inequality:

\begin{equation}
\begin{split}
\log  (1 + \sum_{i \ne {t}} e^{\bm{z}_i-\bm{z}_{max}} ) &\geq \log  (1 + \sum_{i \ne {t}} e^{\bm{z}_{min}-\bm{z}_{max}} ) \\
&=\log (1+(n-1)e^{\bm{z}_{min}-\bm{z}_{max}})
\end{split}
\end{equation}

where $n$ denotes the number of categories. Then we set up a bound $\epsilon$ for cross entropy loss function. We are aware of the necessary condition for cross entropy to be less than or equal to $\epsilon$ is:
\begin{equation}
\label{e8}
   \log (1+(n-1)e^{\bm{z}_{min}-\bm{z}_{max}}) \leq \epsilon 
\end{equation}
so we solve the equation~\ref{e8}:
\begin{equation}
    \bm{z}_{min}-\bm{z}_{max} \geq \log(n-1) - \log(e^{\epsilon}-1)
\end{equation}

As an example:
\begin{equation}
\epsilon = \log 2 \approx 0.69 
\end{equation}
In this case, we are aware that:
\begin{equation}
\log(e^{\epsilon}-1) = 0
\end{equation}
therefore, 
\begin{equation}
\bm{z}_{max}-\bm{z}_{min} \geq \log (n-1) 
\end{equation}

In another word, to make sore the cross entropy loss can be reduced to 0.69, the difference between maximum logit $\bm{z}_{max}$ and minimum logit $\bm{z}_{min}$ must be greater or equal to $\log(n-1)$. However, when it comes to high-dimensional classification problems where $n$ is much greater, $\log(n-1)$ is a relatively massive but unnecessary margin for loss function. 

Therefore, in terms of a classification problem, although we expect that logit of the targeted category is greater than any other non-targeted category, it can result in overfitting problem since conventional cross entropy loss tends to reduce this margin to a large extend, in which it makes the model overlearn the category distribution. However, the margin $\log(n-1)$ is relatively small in sparse-softmax as we diminish the number of category $n$ to hyperparameter $k$, such that alleviate the overfitting problem caused by conventional softmax. 
\begin{table*}
\caption{Experimental results in text classification task}

\centering
\begin{sc}
\begin{small}
\begin{tabular}{ccccccccc}
\hline
 & \multicolumn{2}{c}{\bf WOS-46985} &   \multicolumn{2}{c}{\bf OOS-EVAL} &   \multicolumn{2}{c}{\bf RCV1-V2} &  \multicolumn{2}{c}{\bf IFLYTEK}  \\ 
& Macro F1 & Micro F1 & Macro F1 & Micro F1 & Macro F1 & Micro F1 & Macro F1 & Micro F1 \\ \hline
softmax & 82.15 & 82.65 & 95.53 & 95.19 & 62.04 & 80.52 & 43.29 & 59.5 \\\hline

Sparse(k=1)  & 82.23 & 82.77 & 95.59 & 95.52 & 62.09 & 80.72 & 43.54 & 59.8 \\
Sparse(k=10) & 82.71 & 82.95 & 95.86 & 95.45 & 62.64 & 81.40 & 43.37 & 59.1 \\ 
Sparse(k=20) & \textbf{83.31} & \textbf{83.50} & \textbf{96.08} & \textbf{95.88} & \textbf{63.17} & \textbf{81.52} & 43.85 & \textbf{60.7} \\
Sparse(k=50) & 82.77 & 82.98 & 95.66 & 95.32 & 62.21 & 81.16 & 43.97 & 60.2 \\
Sparse(k=100)& 81.96 & 82.47 & 95.71 & 95.35 & 21.89 & 46.46 & \textbf{44.25} & 60.1 \\

\hline

\end{tabular}
\end{small}
\end{sc}
\label{tc}

\end{table*}

\section{Experiments}

In this section, we compare our sparse-softmax with conventional softmax on several tasks: text classification, abstractive summarization, question generation in the natural language processing field, as well as a image classification task in the computer vision field. Our goal is not to achieve the state-of-the-art on each task but to observe the effect of replacing the original softmax with our sparse-softmax. In the subsection below, we will provide detailed experimental results on the downstream task~\ref{Text Classification}, an efficiency analysis in section~\ref{Efficiency Analysis} and the model performance under different settings of hyperparameter $k$ in section~\ref{Robustness}.

\subsection{Text Classification}
\label{Text Classification}

\paragraph{Datasets. }Text classification is the task of assigning an appropriate category to a given sentence or document. The categories depend on the chosen dataset and can range from topics. In our experiment, we chose 4 different datasets for evaluation, which are all beyond 100 categories. Among them, Web of Science(WOS-46985) dataset~\footnote{https://data.mendeley.com/datasets/9rw3vkcfy4/6} contains 46,985 documents with 134 categories which include 7 parents categories. OOS-eval dataset~\footnote{https://github.com/clinc/oos-eval} is the benchmark for evaluating the 150 types of user intents classification system in the presence of out-of-scope queries for the dialog system. Reuters Corpus Volume I (RCV1-v2)~\footnote{https://archive.ics.uci.edu/ml/datasets/Reuter} consists of more than 800,000 news agency stories manually classified by Reuters Ltd for research purposes, each of which is assigned multiple topics. The total number of topics is 103. IFLYTEK~\footnote{https://global.xfyun.cn/} is a Chinese long text classification dataset, which contains more than 17,000 long text annotated data including various application topics related to daily life, where it has a total of 119 categories. The statistics of these datasets are presented in Table~\ref{dstat}. 

\begin{table*}[h]
\caption{Statistics of datasets in text classification task.}

\centering
\begin{sc}
\begin{small}
\begin{tabular*}{0.62 \hsize}{lrrrr}
\hline
& WoS-46985 & OOS-eval & RCV1-v2 & IFLYTEK  \\\hline

\textbf{training set} \\ 
Samples    & 32889 & 15100 & 775220 & 12133 \\
Categories &  134 & 150 & 103 & 119   \\ 
Average length   &  99 & 8.3 & 120.2 & 289.0  \\
Maximum length &  998 & 28.0 & 500.0 & 4282.0    \\ \hline
\textbf{development set} \\ 
Samples     & 9444 & 3100 & 21510 & 2599 \\
Categories  & 134 & 150 & 103 & 119 \\
Average length    & 200.3 & 8.3 & 120.1 & 289.8 \\
Maximum length  & 1262 & 24.0 & 499.0 & 1755 \\ \hline
\textbf{test set} \\ 
Samples     & 4652 & 5500 & 1191 & 2599 \\
Categories  & 134 & 150 & 103 & 119 \\
Average length    & 197.9 & 8.3 & 116.4 & 289.8 \\
Maximum length  & 691 & 25.0 & 499.0 & 1755 \\ \hline

\end{tabular*}
\end{small}
\end{sc}
\label{dstat}

\end{table*}

\paragraph{Experimental Settings. }

Among on these text classification datasets, except for the Chinese dataset iflytek that we use pre-trained model Nezha\_base~\citep{wei2019nezha} as our baseline,  we all use BERT\_base~\citep{devlin2018bert} as our baseline model on the other English datasets. 
\begin{table*}
\caption{Experimental results on SQuAD 1.1 dataset.}
\label{Squad}
\begin{center}
\begin{sc}
\begin{small}
\begin{tabular*}{0.55 \hsize}{lcccr}
\toprule
& BLEU-1 & BLEU-2 & BLEU-3 & BLEU-4 \\
\midrule
baseline    & 51.49\% & 36.19\% & 27.33\% & 21.20\% \\
sparse-softmax & \textbf{52.39\%} & \textbf{36.81\%} & \textbf{28.13\%} & \textbf{22.02\%}   \\

\bottomrule
\end{tabular*}
\end{small}
\end{sc}
\end{center}
\end{table*}
According to the text length distribution of different datasets, as well as the maximum word length limited by Bert, we set the hyperparameters as shown in the following table~\ref{settings}. It is worth noting that since the text length of the WOS-46985 dataset is generally too long, we adopt the way of head+tail to truncate the information for the text beyond maximum length. Moreover, as the RCV1-V2 dataset contains a large scale of samples, we set epoch is 5 for accelerating the training process, while the rest of the datasets iterate over 20 epochs.

\begin{table}[h]
\caption{Experimental settings in text classification task}

\centering
\begin{sc}
\begin{small}
\begin{tabular}{lrr}
\hline
Dataset & Max length & batch size \\\hline

WOS-46985    & 360 & 16  \\
OOS-EVAL & 64 & 256 \\
RCV1-V2 & 512 & 16 \\ 
IFLYTEK & 256 & 24
\\
\hline

\end{tabular}
\end{small}
\end{sc}
\label{settings}

\end{table}

\paragraph{Results. }
We compare the experimental results of both softmax and sparse-softmax with the setting of K = 1, 10, 20, 50, 100 in four datasets. Following~\citep{johnson2016supervised,howard2018universal}, we adopt macro F1 and mirco F1 score for evaluation and get the following results shown in Table~\ref{tc}. We can observe sparse-softmax all outperforms softmax in four text classification datasets. Empirically, when the number of categories is roughly 100 in certain datasets , the setting of hyperparameter K as 20 tends to achieve the best performance. 
It is worth mentioning that the model performance drops dramatically when we set k as 100 in RCV1-V2 dataset, since the number of categories in the RCV1-V2 dataset is 103 , we found that  the model does not converge properly because of the gradient propagation of the remaining categories in the iterative process when the probability distribution is truncated.

\subsection{Efficiency Analysis} 
\label{Efficiency Analysis}
\begin{figure*}[!ht]
    \centering
    \includegraphics[width = 1.0 \textwidth]{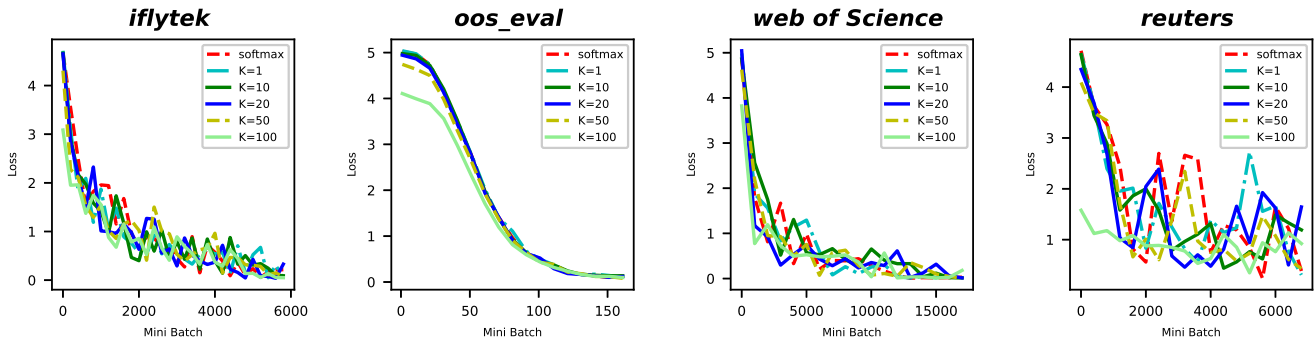}
    \caption{Loss curve in text classification task}
    \label{loss}
\end{figure*}

The plots for training loss upon mini-batches for WoS-46985, OOS-eval, RCV1-v2 and IFLYTEK are given in Figure~\ref{loss}. It can be seen that compared with the traditional softmax, the loss function of sparse-softmax drops to a relatively low level while using fewer epochs when the parameters were consistent with softmax, which proves that sparse-softmax can converge faster in the high-dimensional classification tasks. 
Meanwhile, it may be pertinent to note that in Figure~\ref{loss}, we see ﬂuctuation in the training loss for the softmax function, whereas the plot is comparatively smoother for sparse-softmax, which also indicates our approach can be less likely perturbed and more likely to converge. 

\begin{figure*}[!ht]
    \centering
    \includegraphics[width = 1.0 \textwidth]{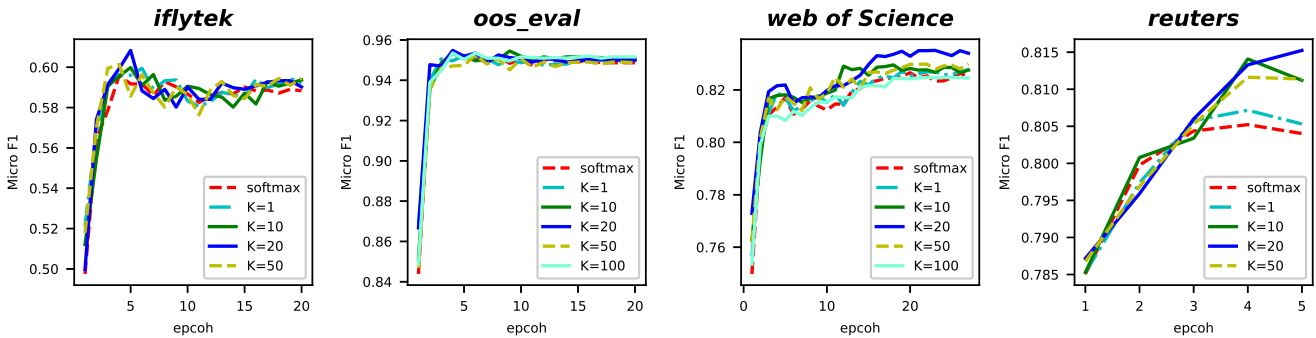}
    \caption{Micro F1 curve in text classification task}
    \label{micro}
\end{figure*}

\subsection{Robustness to hyperparameter $k$} 
\label{Robustness}
In this section, we show that our procedure is stable in its hyperparameter $k$. The 
theoretical results suggest that a wide range of $k$ can give us statistical consistent guarantees of performance improvement against traditional softmax. Such robustness in hyperparameter is highly desirable since optimal tuning is not always feasible under certain circumstances, especially when no sufficient validation set or computational resources are available. 

Empirically, when the number of categories is approximately 100 in the classification tasks , it achieves the state-of-the-art results when $k$ is selected as 20.  As shown in Figure~\ref{micro}, $k = 20$ achieves the best micro F1 scores, which are 60.7\% and 96.1\% in iflytek, oos\_eval dataset respectively.

\section{Auxiliary Experiments}

To verify that sparse-softmax is adaptive across different high-dimensional tasks, we present experiments where we adopt sparse-softmax to the baseline models and compare the model performance respectively. The details are listed in the following subsections including abstractive summarisation and question answering tasks in the natural language processing field, as well as image classification in the computer vision field.

\subsection{Abstractive Summarisation}
Automatic text summarization produces a concise and ﬂuent summary conveying the key information in the input (e.g., a news article). We focus on abstractive summarization, a generation task where the summary is not constrained to reuse the phrases or sentences in the input text. We use Gigaword-10k~\footnote{https://paperswithcode.com/sota/text-summarization-on-gigaword-10k} dataset for evaluation. We also use UNILM~\citep{unilm} as our baseline model and fine-tune the model on training set for 30 epochs. The masking probability is 0.7. We also use label smoothing~\citep{muller2019does} with a rate of 0.1. For Gigaword, we set the batch size to 64, and maximum length to 256. During decoding, we experiment two different beam sizes: 1 and 5, respectively. We also ﬁne-tune UNILM as a sequence-to-sequence model in our task. Due to the massive scale of the input vocabulary, the model needs to classify over 25,731 categories at each time step in the decoding stage if none of the post-processing steps are involved. Even if we mask all the tokens not appearing in the input sentence, the model still needs to classify over 256 categories as we set up the input maximum length to 256. 

We use the F1 version of ROUGE of that in UNILM~\citep{unilm} as the evaluation metric for our dataset. In Table~\ref{summarization}, we compare the model performance of raw baseline with adding our sparse-softmax, we can notice that sparse-softmax outperform baseline by 0.67\% ROUGE-1 and 0.79\% ROUGE-2 when it sets beam size = 1. In terms of beam size = 5, we mask the entire probability distributions except for top 5 ones, and our model also outperforms 0.28\% ROUGE-1 and 0.55\% ROUGE-2 respectively. 

\begin{table}[H]
\caption{Experimental results on Gigaword-10k dataset}

\centering
\begin{tabular}{lrr}
\hline
\textbf{beam size = 1} & ROUGE-1 & ROUGE-2  \\\hline

Unilm + softmax    & 32.23\% & 13.34\% \\
Unilm + sparse-softmax & \textbf{32.90\%} & \textbf{14.13\%}   \\ \hline
\\
\hline
\textbf{beam size = 5} & ROUGE-1 & ROUGE-2  \\\hline
Unilm + softmax    & 32.82\% & 13.93\% \\
Unilm + sparse-softmax & \textbf{33.10\%} & \textbf{14.48\%}   \\ \hline
\end{tabular}

\label{summarization}

\end{table}

\subsection{Question Answering}

We also conduct experiments for the answer-aware question generation task~\citep{chaplot2018gated,lai2017race}. Given an input passage and an answer span, our goal is to generate a question that asks for the answer. We conduct evaluation on The SQuAD 1.1 dataset~\footnote{https://datarepository.wolframcloud.com/resources/SQuAD-v1.1}. The question generation task is also formulated as a sequence-to-sequence problem which means the model needs to classify over the entire vocabulary as same as abstractive summarisation task above.

We also use UNILM as our baseline and ﬁne-tune it on the training set for 10 epochs. We set the batch size to 32, masking probability to 0.7, and learning rate to 2e-5. The rate of label smoothing is 0.1. We set beam size = 1, which is aligned with the baseline model. During decoding, we truncate the input to 464 tokens by selecting a passage chunk that contains the answer. We use BLEU-4 as evaluation metrics~\citep{unilm}. The result is showing in Table~\ref{Squad}, by adding the sparse-softmax, our model outperforms the baseline model by 0.90\%, 0.62\%, 0.80\%, 0.82\% in terms of BLEU-1, BLEU-2, BLEU-3, BLEU-4 respectively. 

\section{Using pre-trained models}

Pretrained model provides more informative initialized parameters to accelerate the convergence speed of the model on specific tasks. However, it also suffers more overfitting problem under the same circumstances as non-pretrained models. As elaborated in Section~\ref{sparse}, we infer that sparse-softmax is capable of alleviating overlearned problem, since it can diminish the gap between maximum logit and minimum logit where it is used to measure the margin of cross entropy loss function. Intuitively, we assume applying sparse-softmax to pretrained model is more effective as the overlearned problem is more obvious in the pretrained models. To verify our speculation and the generalization capability of our approach, we conduct the experiment in an image classification task to verify this hypothesis. We adopt Cifar-100~\footnote{https://www.cs.toronto.edu/~kriz/cifar.html} as our dataset and Densenet 201~\citep{huang2017densely} as our baseline models. We set epoch is 200, batch size is 62, and k is 20 for sparse-softmax in the training stage. The result is shown in Table~\ref{image}, there was a slight performance degradation after using Densenet + Sparse-softmax. 

As we assume that sparse-softmax is only effective under the framework of pre-trained models, we carry out the same experiment with InceptionV3~\citep{szegedy2016rethinking}, which is pre-trained on imageNet dataset~\footnote{http://www.image-net.org/}. After deploy the pre-training framework, our proposed sparse-softmax shows better performance, which verify the hypothesis that sparse-softmax is more suitable for pre-trained model structures.

\begin{table}[H]
\caption{Experimental results on image classification tasks}

\centering
\begin{sc}
\begin{small}
\begin{tabular}{lr}
\hline
Model & Top1 Acc \\\hline

densenet(raw)    & 0.762  \\
densenet(sparse-softmax) & 0.737 \\
InceptionV3(fine tuned) & 0.771 \\ 
InceptionV3(sparse-softmax) & 0.778
\\
\hline

\end{tabular}
\end{small}
\end{sc}
\label{image}

\end{table}

\section{Conclusion}
In this paper, we propose sparse-softmax, which is an alternative of traditional softmax but achieves sparse probability distributions in the output. Experimental results on various tasks verifies that sparse-softmax can convex faster than conventional softmax and gain better model performance in high-dimensional classification tasks. Experiments on the image classification task suggest that our approach is adaptable to different domains under pre-trained model structure.

\bibliography{aaai22}
\clearpage

\appendix

\end{document}